%% file: TheoryDeepDomainAdaptation_main.tex
\documentclass[english]{article}
\usepackage[T1]{fontenc}
\usepackage[latin9]{inputenc}
\usepackage{float}
\usepackage{amsmath}
\usepackage{amsthm}
\usepackage{amssymb}
\usepackage{graphicx}
\usepackage[numbers]{natbib}
\usepackage{xargs}[2008/03/08]
\usepackage{microtype}

\makeatletter

\providecommand{\tabularnewline}{\\}

\theoremstyle{plain}
\newtheorem{thm}{\protect\theoremname}
\theoremstyle{plain}
\newtheorem{prop}[thm]{\protect\propositionname}
\theoremstyle{remark}
\newtheorem{rem}[thm]{\protect\remarkname}
\theoremstyle{plain}
\newtheorem{cor}[thm]{\protect\corollaryname}

\usepackage{graphicx}
\usepackage{subfigure}
\usepackage{booktabs} 
\usepackage{algolyx}

\makeatletter

\usepackage{hyperref}

\usepackage[preprint]{neurips_2019}

\usepackage{algorithm}
\usepackage{algorithmic}
\renewcommand{\algorithmicrequire}{\textbf{Input:}}
\renewcommand{\algorithmicensure}{\textbf{Output:}}

\usepackage{colortbl} 
\definecolor{header_color}{rgb}{0.74,0.88,0.91}
\definecolor{even_color}{rgb}{0.9,0.9,0.9}
\definecolor{subheader_color}{rgb}{0.85,0.93,0.95}
\definecolor{childheader_color}{rgb}{1.0,0.93,0.87}

\definecolor{ccolor_best}{rgb}{1.0,0.9,0.9}
\definecolor{ccolor_wrong}{rgb}{1.0,0.85,0.85}

\usepackage{ifthen}
\renewenvironment{figure}[1][]{%
 \ifthenelse{\equal{#1}{}}{%
   \@float{figure}
 }{%
   \@float{figure}[#1]%
 }%
 \centering
}{%
 \end@float
}

\renewenvironment{table}[1][]{%
 \ifthenelse{\equal{#1}{}}{%
   \@float{table}
 }{%
   \@float{table}[#1]%
 }%
 \centering
}{%
 \end@float
}

\makeatother

\usepackage{babel}
\providecommand{\corollaryname}{Corollary}
\providecommand{\propositionname}{Proposition}
\providecommand{\remarkname}{Remark}
\providecommand{\theoremname}{Theorem}

\begin{document}
\title{On Deep Domain Adaptation: Some Theoretical Understandings}
\author{Trung Le, Khanh Nguyen, Nhat Ho, Hung Bui, Dinh Phung}
\maketitle
\begin{abstract}
Compared with shallow domain adaptation, recent progress in deep domain
adaptation has shown that it can achieve higher predictive performance
and stronger capacity to tackle structural data (e.g., image and sequential
data). The underlying idea of deep domain adaptation is to bridge
the gap between source and target domains in a joint space so that
a supervised classifier trained on labeled source data can be nicely
transferred to the target domain. This idea is certainly intuitive
and powerful, however, limited theoretical understandings have been
developed to support its underpinning principle. In this paper, we
have provided a rigorous framework to explain why it is possible to
close the gap of the target and source domains in the joint space.
More specifically, we first study the loss incurred when performing
transfer learning from the source to the target domain. This provides
a theory that explains and generalizes existing work in deep domain
adaptation which was mainly empirical. This enables us to further
explain why closing the gap in the joint space can directly minimize
the loss incurred for transfer learning between the two domains. To
our knowledge, this offers the first theoretical result that characterizes
a direct bound on the joint space and the gain of transfer learning
via deep domain adaptation.
\end{abstract}
\input{macros.tex}

\section{Introduction\vspace{-2mm}
}

\input{Introduction.tex}

\section{Main Results\vspace{-2mm}
}

\input{TheoResults_NED.tex}

\section{Experiment\vspace{-2mm}
}

\input{Experiment.tex}

\section{Conclusion\vspace{-2mm}
}

\input{Conclusion.tex}

\clearpage{}

\bibliographystyle{plain}

\end{document}

%% file: macros.tex
\newcommand{\sidenote}[1]{\marginpar{\small \emph{\color{Medium}#1}}}

\global\long\def\se{\hat{\text{se}}}%

\global\long\def\interior{\text{int}}%

\global\long\def\boundary{\text{bd}}%

\global\long\def\ML{\textsf{ML}}%

\global\long\def\GML{\mathsf{GML}}%

\global\long\def\HMM{\mathsf{HMM}}%

\global\long\def\support{\text{supp}}%

\global\long\def\new{\text{*}}%

\global\long\def\stir{\text{Stirl}}%

\global\long\def\mA{\mathcal{A}}%

\global\long\def\mB{\mathcal{B}}%

\global\long\def\mF{\mathcal{F}}%

\global\long\def\mK{\mathcal{K}}%

\global\long\def\mH{\mathcal{H}}%

\global\long\def\mX{\mathcal{X}}%

\global\long\def\mZ{\mathcal{Z}}%

\global\long\def\mS{\mathcal{S}}%

\global\long\def\Ical{\mathcal{I}}%

\global\long\def\mT{\mathcal{T}}%

\global\long\def\Pcal{\mathcal{P}}%

\global\long\def\dist{d}%

\global\long\def\HX{\entro\left(X\right)}%
 
\global\long\def\entropyX{\HX}%

\global\long\def\HY{\entro\left(Y\right)}%
 
\global\long\def\entropyY{\HY}%

\global\long\def\HXY{\entro\left(X,Y\right)}%
 
\global\long\def\entropyXY{\HXY}%

\global\long\def\mutualXY{\mutual\left(X;Y\right)}%
 
\global\long\def\mutinfoXY{\mutualXY}%

\global\long\def\given{\mid}%

\global\long\def\gv{\given}%

\global\long\def\goto{\rightarrow}%

\global\long\def\asgoto{\stackrel{a.s.}{\longrightarrow}}%

\global\long\def\pgoto{\stackrel{p}{\longrightarrow}}%

\global\long\def\dgoto{\stackrel{d}{\longrightarrow}}%

\global\long\def\lik{\mathcal{L}}%

\global\long\def\logll{\mathit{l}}%

\global\long\def\vectorize#1{\boldsymbol{#1}}%

\global\long\def\vt#1{\mathbf{#1}}%

\global\long\def\gvt#1{\boldsymbol{#1}}%

\global\long\def\idp{\ \bot\negthickspace\negthickspace\bot\ }%
 
\global\long\def\cdp{\idp}%

\global\long\def\das{\triangleq}%

\global\long\def\id{\mathbb{I}}%

\global\long\def\idarg#1#2{\id\left\{  #1,#2\right\}  }%

\global\long\def\iid{\stackrel{\text{iid}}{\sim}}%

\global\long\def\bzero{\vt 0}%

\global\long\def\bone{\mathbf{1}}%

\global\long\def\boldm{\boldsymbol{m}}%

\global\long\def\be{\boldsymbol{e}}%

\global\long\def\bff{\vt f}%

\global\long\def\ba{\boldsymbol{a}}%

\global\long\def\bb{\boldsymbol{b}}%

\global\long\def\bc{\boldsymbol{c}}%

\global\long\def\bB{\boldsymbol{B}}%

\global\long\def\bx{\boldsymbol{x}}%

\global\long\def\bl{\boldsymbol{l}}%

\global\long\def\bu{\boldsymbol{u}}%

\global\long\def\bo{\boldsymbol{o}}%

\global\long\def\bh{\boldsymbol{h}}%

\global\long\def\bs{\boldsymbol{s}}%

\global\long\def\bz{\boldsymbol{z}}%

\global\long\def\xnew{y}%

\global\long\def\bxnew{\boldsymbol{y}}%

\global\long\def\bX{\boldsymbol{X}}%

\global\long\def\tbx{\tilde{\bx}}%

\global\long\def\by{\boldsymbol{y}}%

\global\long\def\bY{\boldsymbol{Y}}%

\global\long\def\bZ{\boldsymbol{Z}}%

\global\long\def\bU{\boldsymbol{U}}%

\global\long\def\bv{\boldsymbol{v}}%

\global\long\def\bn{\boldsymbol{n}}%

\global\long\def\bV{\boldsymbol{V}}%

\global\long\def\bI{\boldsymbol{I}}%

\global\long\def\bw{\vt w}%

\global\long\def\balpha{\gvt{\alpha}}%

\global\long\def\bbeta{\gvt{\beta}}%

\global\long\def\bmu{\gvt{\mu}}%

\global\long\def\btheta{\boldsymbol{\theta}}%

\global\long\def\bsigma{\boldsymbol{\sigma}}%

\global\long\def\blambda{\boldsymbol{\lambda}}%

\global\long\def\bgamma{\boldsymbol{\gamma}}%

\global\long\def\bpsi{\boldsymbol{\psi}}%

\global\long\def\bphi{\boldsymbol{\phi}}%

\global\long\def\bPhi{\boldsymbol{\Phi}}%

\global\long\def\bpi{\boldsymbol{\pi}}%

\global\long\def\bomega{\boldsymbol{\omega}}%

\global\long\def\bepsilon{\boldsymbol{\epsilon}}%

\global\long\def\btau{\boldsymbol{\tau}}%

\global\long\def\realset{\mathbb{R}}%

\global\long\def\realn{\realset^{n}}%

\global\long\def\integerset{\mathbb{Z}}%

\global\long\def\natset{\integerset}%

\global\long\def\integer{\integerset}%

\global\long\def\natn{\natset^{n}}%

\global\long\def\rational{\mathbb{Q}}%

\global\long\def\rationaln{\rational^{n}}%

\global\long\def\complexset{\mathbb{C}}%

\global\long\def\comp{\complexset}%

\global\long\def\compl#1{#1^{\text{c}}}%

\global\long\def\and{\cap}%

\global\long\def\compn{\comp^{n}}%

\global\long\def\comb#1#2{\left({#1\atop #2}\right) }%

\global\long\def\nchoosek#1#2{\left({#1\atop #2}\right)}%

\global\long\def\param{\vt w}%

\global\long\def\Param{\Theta}%

\global\long\def\meanparam{\gvt{\mu}}%

\global\long\def\Meanparam{\mathcal{M}}%

\global\long\def\meanmap{\mathbf{m}}%

\global\long\def\logpart{A}%

\global\long\def\simplex{\Delta}%

\global\long\def\simplexn{\simplex^{n}}%

\global\long\def\dirproc{\text{DP}}%

\global\long\def\ggproc{\text{GG}}%

\global\long\def\DP{\text{DP}}%

\global\long\def\ndp{\text{nDP}}%

\global\long\def\hdp{\text{HDP}}%

\global\long\def\gempdf{\text{GEM}}%

\global\long\def\Gumbel{\text{Gumbel}}%

\global\long\def\Uniform{\text{Uniform}}%

\global\long\def\Mult{\text{Mult}}%

\global\long\def\rfs{\text{RFS}}%

\global\long\def\bernrfs{\text{BernoulliRFS}}%

\global\long\def\poissrfs{\text{PoissonRFS}}%

\global\long\def\grad{\gradient}%
 
\global\long\def\gradient{\nabla}%

\global\long\def\partdev#1#2{\partialdev{#1}{#2}}%
 
\global\long\def\partialdev#1#2{\frac{\partial#1}{\partial#2}}%

\global\long\def\partddev#1#2{\partialdevdev{#1}{#2}}%
 
\global\long\def\partialdevdev#1#2{\frac{\partial^{2}#1}{\partial#2\partial#2^{\top}}}%

\global\long\def\closure{\text{cl}}%

\global\long\def\cpr#1#2{\Pr\left(#1\ |\ #2\right)}%

\global\long\def\var{\text{Var}}%

\global\long\def\Var#1{\text{Var}\left[#1\right]}%

\global\long\def\cov{\text{Cov}}%

\global\long\def\Cov#1{\cov\left[ #1 \right]}%

\global\long\def\COV#1#2{\underset{#2}{\cov}\left[ #1 \right]}%

\global\long\def\corr{\text{Corr}}%

\global\long\def\sst{\text{T}}%

\global\long\def\SST{\sst}%

\global\long\def\ess{\mathbb{E}}%

\global\long\def\Ess#1{\ess\left[#1\right]}%

\newcommandx\ESS[2][usedefault, addprefix=\global, 1=]{\underset{#2}{\ess}\left[#1\right]}%

\global\long\def\fisher{\mathcal{F}}%

\global\long\def\bfield{\mathcal{B}}%
 
\global\long\def\borel{\mathcal{B}}%

\global\long\def\bernpdf{\text{Bernoulli}}%

\global\long\def\betapdf{\text{Beta}}%

\global\long\def\dirpdf{\text{Dir}}%

\global\long\def\gammapdf{\text{Gamma}}%

\global\long\def\gaussden#1#2{\text{Normal}\left(#1, #2 \right) }%

\global\long\def\gauss{\mathbf{N}}%

\global\long\def\gausspdf#1#2#3{\text{Normal}\left( #1 \lcabra{#2, #3}\right) }%

\global\long\def\multpdf{\text{Mult}}%

\global\long\def\poiss{\text{Pois}}%

\global\long\def\poissonpdf{\text{Poisson}}%

\global\long\def\pgpdf{\text{PG}}%

\global\long\def\wshpdf{\text{Wish}}%

\global\long\def\iwshpdf{\text{InvWish}}%

\global\long\def\nwpdf{\text{NW}}%

\global\long\def\niwpdf{\text{NIW}}%

\global\long\def\studentpdf{\text{Student}}%

\global\long\def\unipdf{\text{Uni}}%

\global\long\def\transp#1{\transpose{#1}}%
 
\global\long\def\transpose#1{#1^{\mathsf{T}}}%

\global\long\def\mgt{\succ}%

\global\long\def\mge{\succeq}%

\global\long\def\idenmat{\mathbf{I}}%

\global\long\def\trace{\mathrm{tr}}%

\global\long\def\argmax#1{\underset{_{#1}}{\text{argmax}} }%

\global\long\def\argmin#1{\underset{_{#1}}{\text{argmin}\ } }%

\global\long\def\diag{\text{diag}}%

\global\long\def\norm{}%

\global\long\def\spn{\text{span}}%

\global\long\def\vtspace{\mathcal{V}}%

\global\long\def\field{\mathcal{F}}%
 
\global\long\def\ffield{\mathcal{F}}%

\global\long\def\inner#1#2{\left\langle #1,#2\right\rangle }%
 
\global\long\def\iprod#1#2{\inner{#1}{#2}}%

\global\long\def\dprod#1#2{#1 \cdot#2}%

\global\long\def\norm#1{\left\Vert #1\right\Vert }%

\global\long\def\entro{\mathbb{H}}%

\global\long\def\entropy{\mathbb{H}}%

\global\long\def\Entro#1{\entro\left[#1\right]}%

\global\long\def\Entropy#1{\Entro{#1}}%

\global\long\def\mutinfo{\mathbb{I}}%

\global\long\def\relH{\mathit{D}}%

\global\long\def\reldiv#1#2{\relH\left(#1||#2\right)}%

\global\long\def\KL{KL}%

\global\long\def\KLdiv#1#2{\KL\left(#1\parallel#2\right)}%
 
\global\long\def\KLdivergence#1#2{\KL\left(#1\ \parallel\ #2\right)}%

\global\long\def\crossH{\mathcal{C}}%
 
\global\long\def\crossentropy{\mathcal{C}}%

\global\long\def\crossHxy#1#2{\crossentropy\left(#1\parallel#2\right)}%

\global\long\def\breg{\text{BD}}%

\global\long\def\lcabra#1{\left|#1\right.}%

\global\long\def\lbra#1{\lcabra{#1}}%

\global\long\def\rcabra#1{\left.#1\right|}%

\global\long\def\rbra#1{\rcabra{#1}}%

%% file: Introduction.tex
Learning a discriminative classifier or other predictor in the presence
of a shift between source (training) and target (testing) distributions
is known as domain adaptation (DA). Domain adaptation aims to devise
automatic methods that make it possible to perform transfer learning
from the source domain with labels to the target domains without labels.
Studies in domain adaptation can be broadly categorized into two themes:
shallow and deep domain adaptations. A number of approaches to domain
adaptation have been suggested in the context of shallow learning
when data representations/features are given and fixed, notably via
reweighing or selecting samples from the source domain \citep{Borgwardt2006,Huang2006,Gong2013}
or seeking an explicit feature space transformation that would map
source distribution into the target ones \citep{Pan2009,Gopalan2011,Baktashmotlagh2013}.

To further advance shallow domain adaptation, deep domain adaptation
has recently been proposed to encourage the learning of new representations
for both source and target data in order to minimize the divergence
between them \citep{Ganin2015,TzengHDS15,long2015,Long:2017:DTL:3305890.3305909,shu2018a,french2018selfensembling}.
Source and target data are mapped to a joint feature space via a generator
and the gap between source and target distributions is bridged in
this joint space by minimizing the divergence between distributions
induced from the source and target domains on this space. For instance,
the works of \citep{Ganin2015,TzengHDS15,long2015,shu2018a,french2018selfensembling}
minimize the Jensen-Shannon divergence between the two relevant distributions
relying on GAN principle \citep{goodfellow2014generative}, while
the work of \citep{long2015} minimizes the maximum mean discrepancy
(MMD) \citep{gretton2007kernel} and the work of \citep{courty2017optimal}
minimizes the Wasserstein distance between them. The idea of bridging
the gap of the source and target domains in a joint feature space
is an intuitive and powerful one. However, to our best of knowledge,
there is limited theoretical work has been proposed to rigorously
explain and provide a theoretical underpinning for this problem.

Some existing work has attempted to study this problem, notably \citep{Mansour2009,Ben-David:2010,redko2017theoretical,zhang19_theory}.
The works of \citep{Ben-David:2010,redko2017theoretical} assumed
the existence of a common hypothesis space used for both source and
target domains and then performed theoretical analysis under this
assumption which might be unrealistic in real-world scenarios. In
particular, the work of \citep{Ben-David:2010} analyzed for the specific
loss function $\left|h(\bx)-f(\bx)\right|$ (i.e., $0-1$ loss) and
its Theorem 1 indicates the target error can be bounded by the sum
of the source error, the total variance distance between the source
and target domains, and the discrepancy between the labeling assignment
distributions of the source and target domains. The work of \citep{redko2017theoretical}
studied for the loss function $\left|h(\bx)-f(\bx)\right|^{q}$, but
its result fails to capture the discrepancy between the labeling assignment
distributions of the source and target domains, which, as shown in
our experiments, is important for the transfer learning performance.

In this paper, we aim to develop a rigorous framework to study the
deep domain adaptation problem. Unlike \citep{Ben-David:2010,redko2017theoretical},
we do not assume that the hypothesis spaces for the source and target
domains are the same. More specifically, the hypothesis spaces for
the target and source domains are shifted by a transformation $T_{ts}$.
Furthermore, our results also hold for any continuous loss function
satisfying a mild condition and is analyzed under a general setting
of having a probabilistic supervisor that assigns labels to data examples
\citep{vapnik99}. Under these much more general conditions than \citep{Ben-David:2010},
we are still able to generalize the result in \citep{Ben-David:2010}
to show that the gap between general losses of two corresponding hypotheses
over the source and target domains is upper-bounded by the Wasserstein
distance between the source distribution and the pushforward distribution
of the target distribution via the transformation $T_{ts}$ plus an
additional discrepancy between the two label assignment distributions
on the two domains.

This result leads us to learn a bijective transformation $T_{ts}$
that minimizes the WS distance between the source distribution and
the induced distribution of the target distribution via $T_{ts}$.
This minimization step further sheds light on the need to close the
gap between the source and target distributions in a joint space and
provides a rigorous underlying explanation for the success in most
of current existing empirical deep domain adaptation work. Moreover,
the theory also indicates that by minimizing the first term (i.e.,
the Wasserstein distance term), we could accidentally increase the
second term (i.e., the discrepancy between two labeling assignment
mechanisms), hence possibly eventually increasing the relevant upper
bound. We conduct the extensive experiments on the synthetic and real-world
datasets to verify the theoretical results obtained and to study the
behaviors when the transport transformation $T_{ts}$ causes the total/partial
match or mismatch of two labeling assignment mechanisms in the joint
space. Last but not least, our result has a strong implication to
unsupervised style transfer (e.g., CycleGAN \citep{zhu2017unpaired}
and DiscoGAN \citep{pmlr-v70-kim17a}) in which one needs to learn
a non-degenerate map that transports the source to target distributions.
Interestingly, our proposed theory can be used to theoretically explain
the formulation of CycleGAN and DiscoGAN, hence it contributes to
deepen the understanding of these popular models in unsupervised style
transfer.

%% file: TheoResults_NED.tex
To facilitate the presentation later, we assume that the data spaces
of the source and target domains are $\mathcal{X}^{s}$ and $\mathcal{X}^{t}$
respectively. Furthermore, we denote the distributions that generate
data samples for the source and target domains as $p^{s}\left(\bx\right)$
(corresponding to the probability measure $\mathbb{P}^{s}$) and $p^{t}\left(\bx\right)$
(corresponding to the probability measure $\mathbb{P}^{t}$) respectively.
We also denote the supervisor distributions that assign labels to
data samples in the source and target domains as $p^{s}\left(y\mid\bx\right)$
and $p^{t}\left(y\mid\bx\right)$ \citep{vapnik99}.

Denote by $\mathcal{H}^{s}:=\left\{ h^{s}:\mathcal{X}^{s}\goto\mathbb{R}\right\} $
the hypothesis set whose elements are used to predict labels source
data. Throughout this paper, we assume that $T_{ts}:\mathcal{X}^{t}\goto\mathbb{\mathcal{X}}^{s}$
is a \textit{bijective} mapping and $T_{st}:=T_{ts}^{-1}$ is the
inverse of $T_{ts}$. Based on the formulation of hypothesis set $\mathcal{H}^{s}$,
we define hypothesis set for target domain as $\mathcal{H}^{t}:=\left\{ h^{t}:\mathcal{X}^{t}\goto\mathbb{R}\mid h^{t}\left(\cdot\right)=h^{s}\left(T_{ts}\left(\cdot\right)\right)\text{ for some \ensuremath{h^{s}\in\mathcal{H}^{s}}}\right\} $.

The intuition behind these definitions and assumptions is that with
$\bx\sim\mathbb{P}^{t}$, we use the mapping $T_{ts}$ to reduce the
difference between two domains and then apply a hypothesis $h^{s}\in\mathcal{H}^{s}$
to predict the label of $\bx$. This gives rise to the question about
the key properties of the transformation mapping $T_{ts}$ so that
we can employ the hypothesis $h^{t}=h^{s}\circ T_{ts}$ to predict
labels of target data where $\circ$ represents the composition function. 

Now, we define by $P^{\#}:=\left(T_{ts}\right)_{\#}\mathbb{P}^{t}$
the pushforward probability distribution induced by transporting $\mathbb{P}^{t}$
via $T_{ts}$. Then, we denote $p^{\#}\left(\bx\right)$ the density
of the probability distribution $P^{\#}$. It induces a new domain,
which is termed as the \textit{transport domain} whose data are generated
from $\mathbb{P}^{\#}$. Given these definitions, we define the supervisor
distribution for the transport domain as $p^{\#}\left(y\mid\bx\right)=p^{t}\left(y\mid\bx^{t}\right)$
where $\bx^{t}=T_{ts}^{-1}\left(\bx\right)=T_{st}\left(\bx\right)$
for any $\bx\in\mathcal{X}^{s}$. To ease the presentation, we denote
the general expected loss as: 
\[
R^{a,b}\left(h\right):=\int\ell\left(y,h\left(\bx\right)\right)p^{b}\left(y\mid\bx\right)p^{a}\left(\bx\right)dyd\bx,
\]
where $a,b$ are in the set $\left\{ s,t,\#\right\} $ and $\ell(\cdot,\cdot)$
specifies a loss function. In addition, we shorten $R^{a,a}$ as $R^{a}$.
Furthermore, given a hypothesis $h^{s}\in\mathcal{H}^{s}$ and $h^{t}=h^{s}\circ T_{ts}$,
we measure the variance of general losses of $h^{s}$ when predicting
on the source domain and general losses of $h^{t}$ when predicting
on the target domain as: 
\begin{align*}
\Delta R\left(h^{s},h^{t}\right):=\left|R^{t}\left(h^{t}\right)-R^{s}\left(h^{s}\right)\right|.
\end{align*}
Finally, for the simplicity of the results in the paper, we consider
solely the case of binary classification where the label $y\in\left\{ -1,1\right\} $.
Please refer to our supplementary material for the relevant background
and the details of all proof.

\subsection{Gap between target and source domains}

\label{subsection:gap_expect_loss} In this subsection, we investigate
the variance $\Delta R\left(h^{s},h^{t}\right)$ between the expected
loss in target domain $R^{t}\left(h^{t}\right)$ and the expected
loss in source domain $R^{s}\left(h^{s}\right)$ where $h^{t}=h^{s}\circ T_{ts}$.
We embark on with the following simple yet key proposition indicating
the connection between $R^{t}\left(h^{t}\right)$ and $R^{\#}\left(h^{s}\right)$. 
\begin{prop}
\label{prop:risk_connections} As long as $h^{t}=h^{s}\circ T_{ts}$,
we have $R^{t}\left(h^{t}\right)=R^{\#}\left(h^{s}\right)$. 
\end{prop}

To derive a relation between $R^{t}\left(h^{t}\right)$ and $R^{s}\left(h^{s}\right)$,
we make the following mild assumption with loss function $\ell$: 
\begin{itemize}
\item[(A.1)] $\sup_{h^{s}\in\mathcal{H}^{s},\bx\in\mathcal{X}^{s},y\in\left\{ -1,1\right\} }\left|\ell\left(y,h^{s}\left(\bx\right)\right)\right|:=M<\infty$. 
\end{itemize}
With simple algebra manipulation, the above assumption is satisfied
when $\ell$ is a bounded loss, e.g., logistic or 0-1 loss or $\ell$
is any continuous loss, $\mathcal{X}^{s}$ is compact, and $\sup_{\bx\in\mathcal{X}^{s}}\left|h^{s}\left(\bx\right)\right|<\infty$.
Equipped with Assumption (A.1), we have the following key result demonstrating
the upper bound of $R^{t}\left(h^{t}\right)$ in terms of $R^{s}\left(h^{s}\right)$.
\begin{thm}
\label{thm:gap_gen_loss} Assume that Assumption (A.1) holds. Then,
for any hypothesis $h^{s}\in\mathcal{H}^{s}$, the following inequality
holds: 
\begin{align*}
\Delta R\left(h^{s},h^{t}\right) & \leq M\left(\text{WS}_{c_{0/1}}\left(\mathbb{P}^{s},\mathbb{P}^{\#}\right)+\min\left\{ \mathbb{E}_{\mathbb{P}^{\#}}\left[\norm{\Delta p\left(y\mid\bx\right)}_{1}\right],\mathbb{E}_{\mathbb{P}^{s}}\left[\norm{\Delta p\left(y\mid\bx\right)}_{1}\right]\right\} \right),
\end{align*}
where $\Delta p\left(y\mid\bx\right)$ is given by $\Delta p\left(y\mid\bx\right):=p^{t}\left(y\mid T_{st}\left(\bx\right)\right)-p^{s}\left(y\mid\bx\right),$
and $WS_{c_{0/1}}\left(\cdot,\cdot\right)$ is the Wasserstein distance
with respect to the cost function $c_{0/1}\left(\bx,\bx'\right)=\mathbf{1}_{\bx\neq\bx'}$,
which returns $1$ if $\bx\neq\bx'$ and $0$ otherwise. 
\end{thm}
\begin{rem}
If the following assumptions hold: 
\end{rem}
\begin{itemize}
\item[(i)] The transformation mapping $T_{st}\left(\bx\right)=\bx$, i.e., we
use the same hypothesis set for both the source and target domains,
\item[(ii)] The loss $\ell\left(y,h\left(\bx\right)\right)=\frac{1}{2}\left|y-h\left(\bx\right)\right|$
where we restrict to consider hypothesis $h:\mathcal{X}\goto\left\{ -1,1\right\} $, 
\end{itemize}
then we recover Theorem 1~in \citep{Ben-David:2010}. 
\begin{rem}
When $\text{WS}_{c_{0/1}}\left(\mathbb{P}^{s},\mathbb{P}^{\#}\right)=0$
(i.e., $\left(T_{ts}\right)_{\#}\mathbb{P}^{t}=\mathbb{P}^{s}$ or
$\left(T_{st}\right)_{\#}\mathbb{P}^{s}=\mathbb{P}^{t}$), and there
is a harmony between two supervisors of source and target domains
(i.e., $p^{s}\left(y\mid\bx\right)=p^{t}\left(y\mid T_{st}\left(\bx\right)\right)$),
Theorem~\ref{thm:gap_gen_loss} suggests that we can do a perfect
transfer learning without loss of performance. This fact is summarized
in the following corollary. 
\end{rem}
\begin{cor}
\label{cor:ideal_case} Assume that $\left(T_{ts}\right)_{\#}\mathbb{P}^{t}=\mathbb{P}^{s}$
(or equivalently $\left(T_{st}\right)_{\#}\mathbb{P}^{s}=\mathbb{P}^{t}$)
and the source and target supervisor distributions are harmonic in
the sense that $p^{s}\left(y\mid\bx\right)=p^{t}\left(y\mid T_{st}\left(\bx\right)\right)$
(or equivalently $p^{t}\left(y\mid\bx\right)=p^{s}\left(y\mid T_{ts}\left(\bx\right)\right)$).
Then, we can do a perfect transfer learning between the source and
target domains. 
\end{cor}

\subsection{Optimization via Wasserstein metric \label{subsec:optimization_Wasserstein}}

Corollary~\ref{cor:ideal_case} suggest that we can do a perfect
transfer learning from the source to target domains if we can point
out a bijective map that transports the target to source distributions
and two supervisor distributions are harmonic via this map. This is
consistent with what is achieved in Theorem \ref{thm:gap_gen_loss}
for which the upper bound of the loss variance $\Delta R\left(h^{s},h^{t}\right)$
vanishes. Particularly, the upper bound in Theorem \ref{thm:gap_gen_loss}
consists of two terms wherein the first term quantifies how distant
the transport and source domains and the second term relates to the
discrepancy of two supervisor distributions. Still, from Theorem \ref{thm:gap_gen_loss},
we obtain the following inequality:
\[
R^{t}\left(h_{t}\right)\leq R^{s}\left(h_{s}\right)+M\left(\text{WS}_{c_{0/1}}\left(\mathbb{P}^{s},\mathbb{P}^{\#}\right)+\min\left\{ \mathbb{E}_{\mathbb{P}^{\#}}\left[\norm{\Delta p\left(y\mid\bx\right)}_{1}\right],\mathbb{E}_{\mathbb{P}^{s}}\left[\norm{\Delta p\left(y\mid\bx\right)}_{1}\right]\right\} \right),
\]
which requires us to find the best hypothesis $h_{s}^{*}$ and transformation
$T_{ts}^{*}$ for minimizing the general loss $R^{s}\left(h_{s}\right)$
and the remaining term. To minimize the remaining term, due to the
lack of target labels, it is natural to focus on minimizing the first
term $\text{WS}_{c_{0-1}}\left(\mathbb{P}^{s},\mathbb{P}^{\#}\right)$
by restricting the transformation $T_{ts}$ in the family of those
what can transport the target to source distributions. By this restriction,
the problem of interest boils down to answering the question: \emph{among
the bijective maps $T_{ts}$ that transport the target to source distributions
which transformation incurs the minimal discrepancy as specified in
the second term of the upper bound in Theorem \ref{thm:gap_gen_loss}}.

We further tackle the task of finding the bijective maps that transport
the target to source distributions via the Wasserstein distance with
respect to the cost function (or metric) $c$ and $p>0$ as:
\begin{equation}
\min_{H}\,WS_{c,p}\left(H_{\#}\mathbb{P}^{t},\mathbb{P}^{s}\right),\label{eq:ws_op}
\end{equation}
where $WS_{c,p}\left(\mathbb{P},\mathbb{Q}\right)=\inf_{T_{\#}\mathbb{P}=\mathbb{Q}}\mathbb{E}_{\bx\sim\mathbb{P}}\left[c\left(\bx,T\left(\bx\right)\right)^{p}\right]^{1/p}$
is a Wasserstein distance between two distributions $\mathbb{P}$
and $\mathbb{Q}$.

\begin{figure}
\begin{centering}
\vspace{-2mm}
\includegraphics[width=0.6\columnwidth]{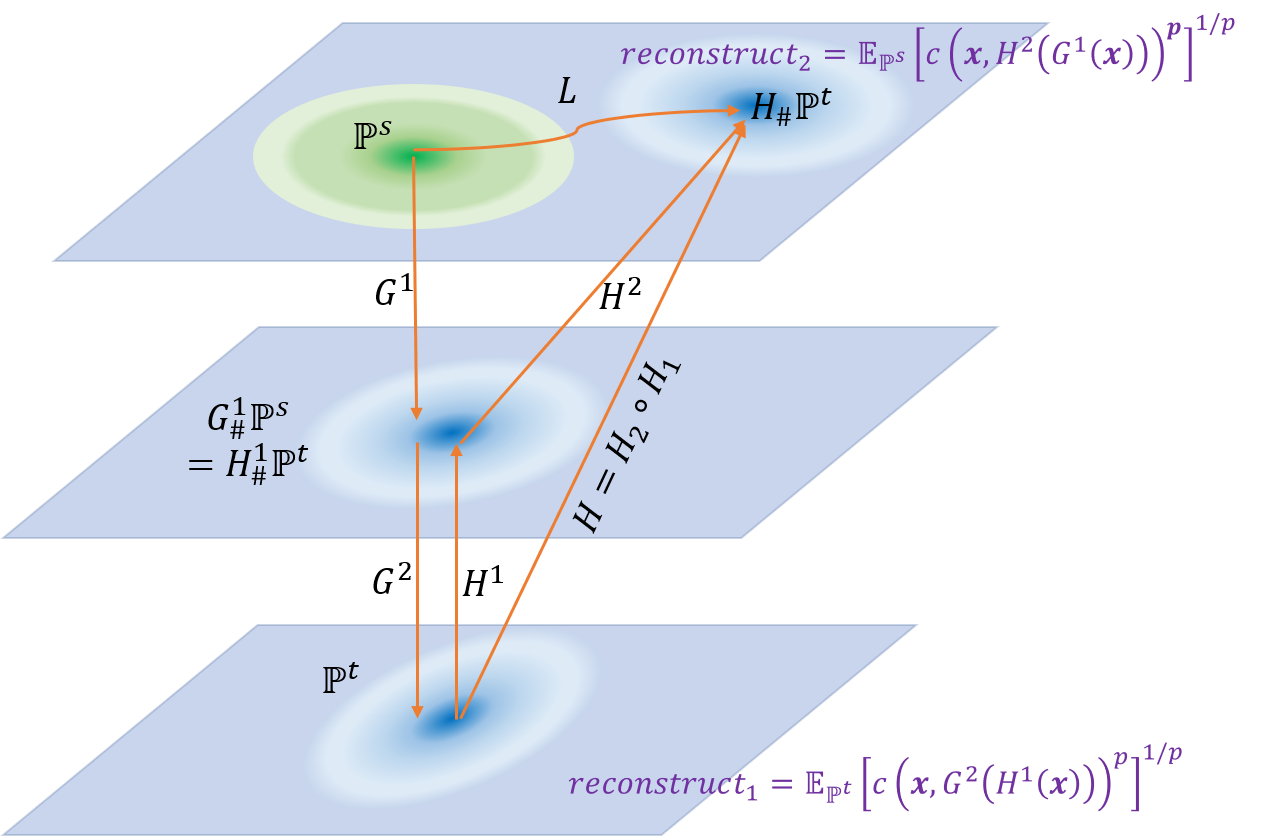} \vspace{-3mm}
\par\end{centering}
\caption{Theoretical view of Deep Domain Adaptation. The mapping $H=H^{2}\circ H^{1}$
maps from the target to source domains, while the mapping $G=G^{2}\circ G^{1}$
maps from the source to target domains. We minimize $D\left(G_{\#}^{1}\mathbb{P}^{s},H_{\#}^{1}\mathbb{P}^{t}\right)$
to close the discrepancy gap of the source and target domains in the
joint space. In addition, we further minimize the reconstruction terms
to avoid the mode collapse. Interestingly, our formulation has a strong
connection to unsupervised style transfer (e.g., CycleGAN \citep{zhu2017unpaired},
DiscoGAN \citep{pmlr-v70-kim17a}). \label{fig:theo_view}}
\vspace{-5mm}
\end{figure}

Let $\mathcal{Z}$ be an intermediate space (i.e., the joint space
$\mathcal{Z}=\mathbb{R}^{m}$). We consider the composite mappings
$H$: $H\left(\bx\right)=H^{2}\left(H^{1}\left(\bx\right)\right)$
where $H^{1}$ is an injective mapping from the target domain $\mathcal{X}^{t}$
to the joint space $\mathcal{Z}$ and $H^{2}$ maps from the joint
space $\mathcal{Z}$ to the source domain $\mathcal{X}^{s}$ (note
that if $\mathcal{Z}=\mathcal{X}^{s}$ then $H^{2}=id$ is the identity
function). Based on that structure on $H$, we can recast the optimization
with Wasserstein metric in~\eqref{eq:ws_op} into the following optimization
problem: 
\begin{align}
\min_{H^{1},H^{2}}\,\text{WS}_{c,p}\left(\left(H^{2}\circ H^{1}\right)_{\#}\mathbb{P}^{t},\mathbb{P}^{s}\right).\label{eq:optimization_Wasserstein_neural}
\end{align}
In the following theorem, we demonstrate that the above optimization
problem can be equivalently transformed into another form involving
the joint space (see Figure \ref{fig:theo_view} for an illustration
of that theorem). 
\begin{thm}
\label{thm:equi_wasserstein_form} The optimization problem~\eqref{eq:optimization_Wasserstein_neural}
is equivalent to the following optimization problem: 
\begin{align}
\min_{H^{1},H^{2}}\min_{G^{1}:H_{\#}^{1}\mathbb{P}^{t}=G_{\#}^{1}\mathbb{P}^{s}}\,\mathbb{E}_{\bx\sim\mathbb{P}^{s}}\left[c\left(\bx,H^{2}\left(G^{1}\left(\bx\right)\right)\right)^{p}\right]^{1/p},\label{eq:optimization_Wasserstein_neural_equi}
\end{align}
where $G^{1}$ is an injective map from the source domain $\mathcal{\mathcal{X}}^{s}$
to the joint space $\mathcal{Z}$.
\end{thm}
It is interesting to interpret $G^{1}$ and $H^{1}$ as two generators
that map the source and target domains to the common joint space $\mathcal{Z}$
respectively. The constraint $H_{\#}^{1}\mathbb{P}^{t}=G_{\#}^{1}\mathbb{P}^{s}$
further indicates that the gap between the source and target distributions
is closed in the joint space via two generators $G^{1}$ and $H^{1}$.
Furthermore, $H^{2}$ maps from the joint space to the source domain
and aims to reconstruct $G^{1}$, hence ensuring $G^{1}$ to be injective.
Similar to \citep{Tolstikhin2018WassersteinA}, we do relaxation and
arrive at the optimization problem:
\begin{equation}
\min{}_{H^{1},H^{2},G^{1}}\left(\,\mathbb{E}_{\bx\sim\mathbb{P}^{s}}\left[c\left(\bx,H^{2}\left(G^{1}\left(\bx\right)\right)\right)^{p}\right]^{1/p}+\alpha D\left(G_{\#}^{1}\mathbb{P}^{s},H_{\#}^{1}\mathbb{P}^{t}\right)\right),\label{eq:ws_relax}
\end{equation}
where $D\left(\cdot,\cdot\right)$ specifies a divergence between
two distributions over the joint space and $\alpha>0$.

It is obvious that when the trade-off parameter $\alpha$ approaches
$+\infty$, the solution of the relaxation problem in Eq. \eqref{eq:ws_relax}
approaches the optimal solution in Eq. \eqref{eq:optimization_Wasserstein_neural_equi}.
Moreover, we add another reconstruction term to ensure the injection
of $H^{1}$ and come with the optimization problem:{\small{}
\begin{gather}
\min{}_{H^{1:2},G^{1:2}}\,\biggl(\mathbb{E}_{\mathbb{P}^{s}}\left[c\left(\bx,H^{2}\left(G^{1}\left(\bx\right)\right)\right)^{p}\right]^{1/p}+\mathbb{E}_{\mathbb{P}^{t}}\left[c\left(\bx,G^{2}\left(H^{1}\left(\bx\right)\right)\right)^{p}\right]^{1/p}+\alpha D\left(G_{\#}^{1}\mathbb{P}^{s},H_{\#}^{1}\mathbb{P}^{t}\right)\biggr).\label{eq:ws_recon}
\end{gather}
}{\small\par}

To enable the transfer learning, we can train a supervised classifier
$\mathcal{C}$ on either $\mathcal{D}^{s}=\left\{ \left(\bx_{1}^{s},y_{1}\right),\dots,\left(\bx_{N_{s}}^{s},y_{s}\right)\right\} $
or $G^{1}\left(\mathcal{D}^{s}\right)=\left\{ \left(G^{1}\left(\bx_{1}^{s}\right),y_{1}\right),\dots,\left(G^{1}\left(\bx_{N_{s}}^{s}\right),y_{s}\right)\right\} $.
The final optimization problem is hence as follows: 
\begin{gather}
\min{}_{H^{1:2},G^{1:2}}\biggl(\mathbb{E}_{\bx\sim\mathbb{P}^{s}}\left[c\left(\bx,H^{2}\left(G^{1}\left(\bx\right)\right)\right)^{p}\right]^{1/p}+\mathbb{E}_{\bx\sim\mathbb{P}^{t}}\left[c\left(\bx,G^{2}\left(H^{1}\left(\bx\right)\right)\right)^{p}\right]^{1/p}\nonumber \\
\,\,\,\,\,\,\,\,\,\,\,\,\,\,\,\,\,\,\,\,\,\,\,\,\,\,\,\,\,\,\,\,\,\,\,\,\,\,\,\,\,\,\,\,\,\,\,\,\,\,\,\,\,\,\,\,\,\,\,\,\,\,\,\,\,\,\,\,\,+\alpha D\left(G_{\#}^{1}\mathbb{P}^{s},H_{\#}^{1}\mathbb{P}^{t}\right)+\beta\mathbb{E}_{\left(\bx,y\right)\sim\mathcal{D}^{s}}\left[\ell\left(y,\mathcal{C}\left(A\left(\bx\right)\right)\right)\right]\biggr),\label{eq:op_ws_clf}
\end{gather}
where $A$ is either the identity map (if we train the classifier
$\mathcal{C}$ on $\mathcal{D}^{s}$) or $G^{1}$ (if we train the
classifier $\mathcal{C}$ on $G^{1}\left(\mathcal{D}^{s}\right)$)
and $\beta>0$. 

Since $G^{1}$ and $H^{1}$ are two injective maps from the source
and target domains to the joint space, we can further define two source
and target supervisor distributions on the joint space as $p^{\#,s}\left(y\mid G^{1}\left(\bx\right)\right)=p^{s}\left(y\mid\bx\right)$
and $p^{\#,t}\left(y\mid H^{1}\left(\bx\right)\right)=p^{t}\left(y\mid\bx\right)$.
With respect to the joint space, the second term of the upper bound
in Theorem \ref{thm:gap_gen_loss} can be rewritten as in the following
corollary. 
\begin{cor}
\label{cor:joint_second}The second term of the upper bound in Theorem
\ref{thm:gap_gen_loss} can be rewritten as{\small{}
\[
\min\left\{ \mathbb{E}_{\mathbb{P}^{s}}\left[\norm{p^{\#,t}\left(y\mid G^{1}\left(\bx\right)\right)-p^{\#,s}\left(y\mid G^{1}\left(\bx\right)\right)}_{1},\mathbb{E}_{\mathbb{P}^{t}}\left[\norm{p^{\#,t}\left(y\mid H^{1}\left(\bx\right)\right)-p^{\#,s}\left(y\mid H^{1}\left(\bx\right)\right)}_{1}\right]\right]\right\} .
\]
}{\small\par}
\end{cor}
It is worth noting that solving the optimization problem \eqref{eq:op_ws_clf}
is only a part of the problem of interest since although the optimal
map $H=H^{2}\circ H^{1}$ transports the target to source distributions,
the discrepancy gap in labeling source and target domains is likely
high. Moreover, Corollary \ref{cor:joint_second} sheds light on when
this discrepancy gap in labeling (the second term of the upper bound)
is low or high, that is, this gap is low if $G^{1}$ and $H^{1}$
map the corresponding classes of the source and target domains together
in the joint space and in contrast, this gap is high if there is a
mismatch as mapping the corresponding classes to the joint space.
In the experimental section, we design experiments to demonstrate
these behaviors and how the harmony in these two labeling assignment
mechanisms affects the predictive performance. In addition, two reconstruction
terms in \eqref{eq:op_ws_clf} contribute to preserve the clustering
structures of source and target domains in the joint space, hence
helps to reduce the mode collapsing problem. Finally, in deep domain
adaption, we employ multilayered neural networks to formulate the
transporting transformation $H$ and the joint space specifies an
intermediate layer in this this network.

\paragraph{Further discussion. }

Our formulation in Eq. \eqref{eq:ws_recon} has a strong implication
to unsupervised style transfer (e.g., CycleGAN \citep{zhu2017unpaired}
and DiscoGAN \citep{pmlr-v70-kim17a}) wherein one needs to learn
a non-degenerate map that transports the source to target distributions.
Our theory can mathematically explain what has been done in CycleGAN
and DiscoGAN. In particular, the common formulation in CycleGAN \citep{zhu2017unpaired}
and DiscoGAN \citep{pmlr-v70-kim17a} is in the spectrum of our \eqref{eq:ws_recon}
as we consider the L1 metric: $c\left(\bx,\bx'\right)=\norm{\bx-\bx'}_{1}$
and set $p=1$, and the joint (intermediate) layer is the last layer
of $H$ (i.e., $H^{1}(\bx)=H(\bx),\,H^{2}\left(\bx\right)=\bx$) and
the first layer $G$ (i.e., $G^{1}\left(\bx\right)=\bx,\,G^{2}\left(\bx\right)=G\left(\bx\right)$).
With this setting, our Eq. \eqref{eq:ws_recon} reads:
\begin{equation}
\min{}_{H,G}\,\biggl(\mathbb{E}_{\mathbb{P}^{t}}\left[\norm{\bx-G\left(H\left(\bx\right)\right)}_{1}\right]+\alpha D\left(\mathbb{P}^{s},H_{\#}\mathbb{P}^{t}\right)\biggr).\label{eq:reduce_ws}
\end{equation}

Compared with the formulation in CycleGAN \citep{zhu2017unpaired}
and DiscoGAN \citep{pmlr-v70-kim17a}, Eq. \eqref{eq:reduce_ws} lacks
of the opposite term: $\mathbb{E}_{\mathbb{P}^{s}}\left[\norm{\bx-H\left(G\left(\bx\right)\right)}_{1}\right]+\alpha D\left(\mathbb{P}^{t},G_{\#}\mathbb{P}^{s}\right)$.
However, this missing term can be introduced in our theory if we learn
another transformation $G$ that transports $\mathbb{P}^{s}$ to $\mathbb{P}^{t}$
via a Wasserstein distance.

%% file: Experiment.tex
We conduct the experiments on the synthetic dataset to empirically
verify our proposed theory and on the real-world datasets to demonstrate
the influence of reconstruction terms and how harmony in two labeling
assignment mechanisms affects the predictive performance. To bridge
the gap between the source and target domains, inspired by \citep{Ganin2015},
we employ  GAN principle \citep{goodfellow2014generative}.\vspace{-1mm}

\subsection{Experiment on Synthetic Data\vspace{-1mm}
}

\subsubsection{Synthetic Dataset for the Source and Target Domains}

\label{subsec:exp_synthetic} We generate two synthetic labeled datasets
for the source and target domains. We generate the $10,000$ data
examples of the source dataset from the mixture of two Gaussian distributions:
$p^{s}\left(\bx\right)=\pi_{1}^{s}\mathcal{N}\left(\bx\mid\bmu_{1}^{s},\Sigma_{1}^{s}\right)+\pi_{2}^{s}\mathcal{N}\left(\bx\mid\bmu_{2}^{s},\Sigma_{2}^{s}\right)$
where $\pi_{1}^{s}=\pi_{2}^{s}=\frac{1}{2}$, $\bmu_{1}^{s}=\left[1,1,...,1\right]\in\mathbb{R}^{10},\,\bmu_{2}^{s}=\left[2,2,...,2\right]\in\mathbb{R}^{10}$
and $\Sigma_{1}^{s}=\Sigma_{2}^{s}=\mathbb{I}_{10}$. Similarly, we
generate the another $10,000$ data examples of the target dataset
from the mixture of two Gaussian distributions: $p^{t}\left(\bx\right)=\pi_{1}^{t}\mathcal{N}\left(\bx\mid\bmu_{1}^{t},\Sigma_{1}^{t}\right)+\pi_{2}^{t}\mathcal{N}\left(\bx\mid\bmu_{2}^{t},\Sigma_{2}^{t}\right)$
where $\pi_{1}^{t}=\frac{1}{3}$, $\pi_{2}^{t}=\frac{2}{3}$, $\bmu_{1}^{t}=\left[4,4,...,4\right]\in\mathbb{R}^{10},\,\bmu_{2}^{t}=\left[5,5,...,5\right]\in\mathbb{R}^{10}$
and $\Sigma_{1}^{t}=\Sigma_{2}^{t}=\mathbb{I}_{10}$. For each data
example in the source and target domains, we assign label $y=0$ if
this data example is generated from the first Gauss and $y=1$ if
this data example is generated from the second Gauss using Bayes'
s rule. 
\begin{figure}[H]
\centering{}\vspace{-2mm}
 \includegraphics[width=0.45\textwidth]{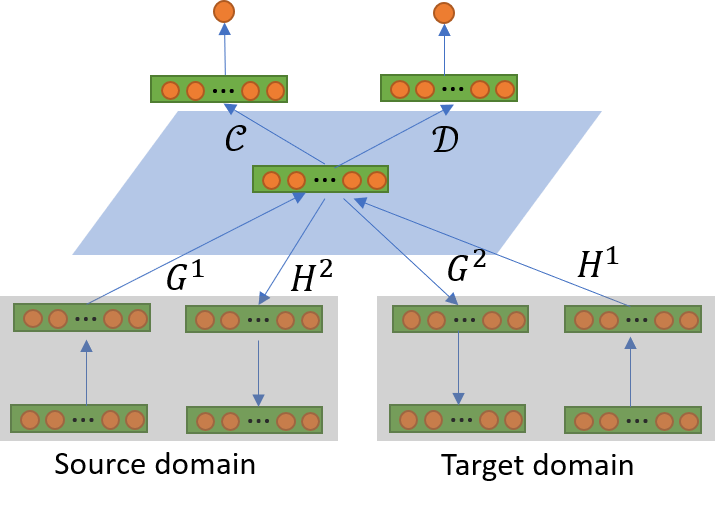}\vspace{-5mm}
 \caption{Architecture of networks for deep domain adaptation.\label{fig:Architecture}}
\vspace{-4mm}
 
\end{figure}

\subsubsection{Deep Domain Adaptation on the Synthetic Dataset\vspace{-1mm}
}

Figure \ref{fig:Architecture} shows the architectures of networks
used in our experiments on the synthetic datasets. Two generators
$G^{1},H^{1}$ with the same architectures ($10\goto5\,(\text{ReLu})\goto5\,(\text{ReLu})$)
map the source and target data to the intermediate joint layer. Note
that different from other works in deep domain adaptation, we did
not tie $G^{1}$ and $H^{1}$. Two other networks $G^{2},H^{2}$ with
the same architectures ($10\goto5\,(\text{ReLu})\goto5\,(\text{ReLu})$)
map from the intermediate joint layer to the source and target domains
respectively. The purpose of $G^{2},H^{2}$ is to reconstruct $H^{1},G^{1}$
respectively. To break the gap between the source and target domains
in the joint layer, we employ GAN principle \citep{goodfellow2014generative,Ganin2015}
wherein we invoke a discriminator network $\mathcal{D}$ ($5\goto5\,(\text{ReLu})\goto1\,(\text{sigmoid})$)
to discriminate the source and target data examples in the joint space.
The classifier network $\mathcal{C}$ ($5\goto5\,(\text{ReLu})\goto1\,(\text{sigmoid})$)
is employed to classify the labeled source data examples. To approximate
the $0/1$ cost function, we use the modified sigmoid function \citep{pmlr-v28-nguyen13a}:
$c_{\gamma}\left(\bx,\bx'\right)=2/\left[1+\exp\left\{ -\gamma\norm{\bx-\bx'}_{2}\right\} \right]-1$
with $\gamma=100$. It can be seen that when $\gamma\goto+\infty$,
the cost function $c_{\gamma}$ approaches the $0/1$ cost function.
More specifically, we need to update $G^{1:2},H^{1:2},\mathcal{C}$,
and $\mathcal{D}$ as follows: 
\begin{align*}
\left(G^{1:2},H^{1:2},\mathcal{C}\right) & =\argmin{G^{1:2},H^{1:2},\mathcal{C}}\mathcal{I}\left(G^{1:2},H^{1:2},\mathcal{C}\right)\text{ and \ensuremath{\mathcal{D}=}\ensuremath{\argmax{\mathcal{D}}\,}\ensuremath{\mathcal{J}\left(\mathcal{D}\right)}},
\end{align*}
where{\small{} $\alpha$ is set to $0.1$ and} we have defined{\small{}
\begin{gather*}
\mathcal{I}\left(G^{1:2},H^{1:2},\mathcal{C}\right)=\mathbb{E}_{\bx\sim\mathbb{P}^{t}}\left[c_{\gamma}\left(\bx,G^{2}\left(H^{1}\left(\bx\right)\right)\right)\right]\\
+\mathbb{E}_{\bx\sim\mathbb{P}^{s}}\left[c_{\gamma}\left(\bx,H^{2}\left(G^{1}\left(\bx\right)\right)\right)\right]+\mathbb{E}_{\left(\bx,y\right)\sim\mathcal{D}^{s}}\left[\ell\left(y,\mathcal{C}\left(G\left(\bx\right)\right)\right)\right]\\
+\alpha\left[\mathbb{E}_{\bx\sim\mathbb{P}^{s}}\left[\log\left(\mathcal{D}\left(G\left(\bx\right)\right)\right)\right]+\mathbb{E}_{\bx\sim\mathbb{P}^{t}}\left[\log\left(1-\mathcal{D}\left(G\left(\bx\right)\right)\right)\right]\right]\\
\mathcal{J}\left(\mathcal{D}\right)=\mathbb{E}_{\bx\sim\mathbb{P}^{s}}\left[\log\left(\mathcal{D}\left(G\left(\bx\right)\right)\right)\right]+\mathbb{E}_{\bx\sim\mathbb{P}^{t}}\left[\log\left(1-\mathcal{D}\left(G\left(\bx\right)\right)\right)\right].
\end{gather*}
}{\small\par}

Based on the classifier $\mathcal{C}$ on the joint space, we can
identify the corresponding hypotheses on the source and target domains
as: $h^{s}\left(\bx\right)=\mathcal{C}\left(G\left(\bx\right)\right)$
and $h^{t}\left(\bx\right)=\mathcal{C}\left(H\left(\bx\right)\right)$.\vspace{-1mm}

\subsubsection{Verification of Our Theory for Unsupervised Domain Adaptation\vspace{-1mm}
}

\begin{figure}[H]
\begin{centering}
\vspace{-3mm}
 \includegraphics[width=0.95\textwidth]{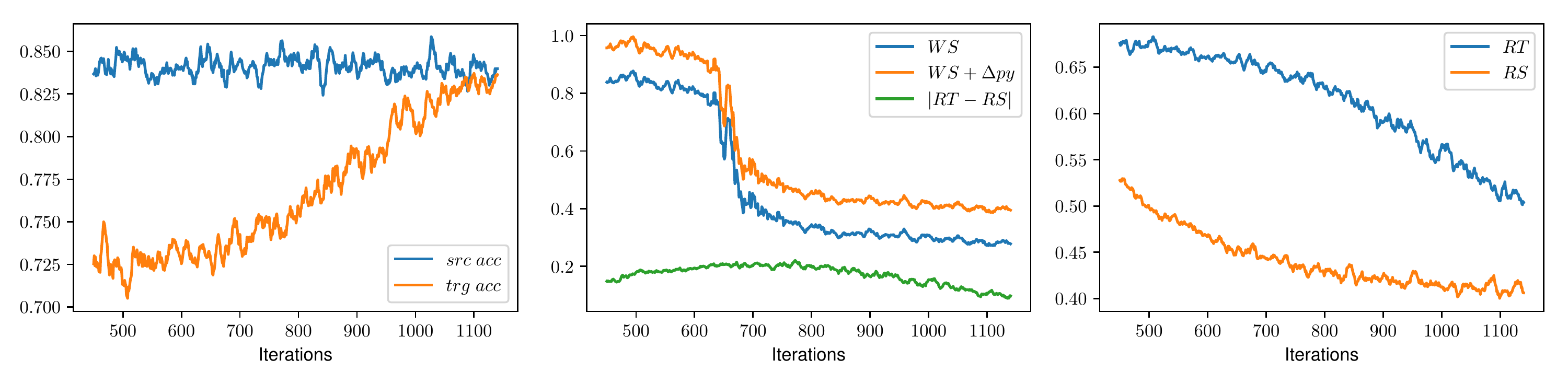}\vspace{-5mm}
\par\end{centering}
\caption{Left: the accuracies on the source and target datasets. Middle: the
plots of three terms in Theorem \ref{thm:gap_gen_loss}. Right: the
plot of empirical losses on the source and target datasets.\label{fig:unsupervised}}
\vspace{-2mm}
\end{figure}

In this experiment, we assume that none of data example in the target
domain has label. We measure three terms, namely $\left|R\left(h^{t}\right)-R\left(h^{s}\right)\right|,\,WS\left(\mathbb{P}^{s},\mathbb{P}^{\#}\right)$,
and $\min\left\{ \mathbb{E}_{\mathbb{P}^{\#}}\left[\norm{\Delta p\left(y\mid\bx\right)}_{1}\right],\mathbb{E}_{\mathbb{P}^{s}}\left[\norm{\Delta p\left(y\mid\bx\right)}_{1}\right]\right\} $
($M=1$ since we are using the logistic loss) as defined in Theorem
\ref{thm:gap_gen_loss} across the training progress. Actually, we
approximate $R\left(h^{t}\right),R\left(h^{s}\right)$ using the corresponding
empirical losses. As shown in Figure \ref{fig:unsupervised} (middle),
the green plot is always above the blue plot and this empirically
confirms the inequality in Theorem \ref{thm:gap_gen_loss}. Furthermore,
the fact that three terms consistently decrease across the training
progress indicates an improvement when $\mathbb{P}^{\#}$ is shifting
toward $\mathbb{P}^{s}$. This improvement is also reflected in Figure
\ref{fig:unsupervised} (left and right) wherein the target accuracy
and empirical loss gradually increase and decrease accordingly.

\subsection{Experiment on Real-world Datasets}

We conduct the experiments on the real-world datasets to demonstrate
the effect of the reconstruction term to the predictive performance
and also the behaviors when the transport transformation $T_{ts}$
causes the match or mismatch of two labeling assignment mechanisms
in the joint space (i.e., we properly and improperly align the classes
of two domains in the joint space). It is worth noting that we do
not seek the state-of-the-art performance in our experiments. Alternatively,
we only focus on investigating the behaviors of the additional theoretical
components. Bearing this in mind, we base on the relevant architectures
and experimental protocols in \citep{Ganin2015} and then start adding
the additional components.\vspace{-1mm}

\subsubsection{The Effect of The Reconstruction Term\vspace{-1mm}
}

As mentioned before, we make use of the relevant architectures for
the generators $G^{1},H^{1}$, the discriminator $\mathcal{D}$, and
the domain classifier $\mathcal{C}$ as proposed in \citep{Ganin2015}
and then add additional components $G^{2},H^{2}$ for the reconstruction
terms. Note that unlike \citep{Ganin2015} and other works, we do
not tie the parameters of two generators $G^{1}$ and $H^{1}$. We
always tie the parameters of $G^{1},H^{2}$ and $H^{1},G^{2}$ because
they are encoders and decoders which aim to reconstruct source and
target samples respectively. Moreover, we slightly modify Eq. \eqref{eq:op_ws_clf}
by setting the hyper-parameter $\beta$ to $1$ and introducing $\theta>0$
as the hyper-parameter of the reconstruction terms. The hyper-parameter
$\alpha$ (corresponding to the adaptation factor $\lambda$ in \citep{Ganin2015})
is scheduled as proposed in that paper. Finally, we search $\theta$
in the grid $\left\{ 0.2,0.4,0.6,0.8,1\right\} $ and obtain the following
experimental results.

\begin{table}[h]
\begin{centering}
\vspace{-2mm}
\begin{tabular}{|c|c|c|c|c|c|c|}
\cline{2-7} 
\multicolumn{1}{c|}{} & \multicolumn{5}{c|}{\textbf{Our Model}} & \textbf{DANN}\citep{Ganin2015}\tabularnewline
\hline 
$\boldsymbol{\theta}$ & $\theta=0.2$ & $\theta=0.4$ & $\theta=0.6$ & $\theta=0.8$ & $\theta=1.0$ & $\theta=0$\tabularnewline
\hline 
\textbf{MNIST$\goto$MNIST-M} & 81.7 & 83.2 & \textbf{88.7} & 85.6 & 80.2 & 81.5\tabularnewline
\hline 
\textbf{SVHN$\goto$MNIST} & 68.4 & 67.8 & 64.4 & 62.8 & 61.3 & \textbf{71.0}\tabularnewline
\hline 
\end{tabular}\caption{The variation of predictive performance in percentage when adding
the reconstruction terms. Note that as $\theta=0$, our model coincides
DANN in \citep{Ganin2015}. We emphasize in bold the best performance.
\label{tab:reconstruct}}
\vspace{-5mm}
\par\end{centering}
\end{table}
In the experimental results in Table \eqref{tab:reconstruct}, for
the first pair, the predictive performance starts increasing, peaks
at its maximum, and then drops, whereas the predictive performance
for the second pair peaks at $\theta=0$ (i.e., no reconstruction
term). This shows that preserving the geometry/cluster structure of
the source/target domains in the joint space is an ingredient that
helps improve the predictive performance and avoid the mode collapse,
but might cause the source and target examples harder to mix up in
the joint space because this adds more constraint to this process
(e.g., there is no improvement for the second pair). In practice,
we can tweak this corresponding trade-off parameter to partly preserving
the geometry/cluster structures in the original spaces whereas making
it convenient for mixing up the source and target examples in the
joint space.\vspace{-1mm}

\subsubsection{The Effect of Class Alignment in the Joint Space\vspace{-1mm}
}

In this experiment, we inspect the influence of the harmony of two
labeling assignment mechanisms to the predictive performance. In particular,
we assume that a portion ($r=5\%,\,15\%,\,25\%,\,50\%$ ) of the target
domain has label and consider two settings: i) the labels of the target
and source domains are totally properly matched in the joint space
(i.e., $0$ matches $0$, $1$ matches $1$,..., and $9$ matches
$9$) and ii) the labels of the target and source domains are totally
improperly matches in the joint space (i.e., $0$ matches $1$, $1$
matches $2$,..., and $9$ matches $0$). 

To push a specific labeled portion of the target domain to the corresponding
label portion of the source domain in the joint space (the label $i$
to $i$ in the first setting and the label $i$ to $(i+1)\,\text{mod}\,10$
in the second setting for $i=0,1,\dots,9$), we again make use of
GAN principle and employ additional discriminators to push the corresponding
labeled portions together. Note that the parameters of the additional
discriminators and the primary discriminator (used to push the target
data toward source data in the joint space) are tied up to the penultimate
layer. 

\begin{table}[h]
\centering{}\vspace{-2mm}
\begin{tabular}{|c|c|c|c|c|c|c|c|c|c|}
\cline{2-10} 
\multicolumn{1}{c|}{} & \multicolumn{4}{c|}{\textbf{Proper match}} & \multicolumn{4}{c|}{\textbf{Improper match}} & \textbf{Base}\tabularnewline
\hline 
$\boldsymbol{r}$ & $5\%$ & $15\%$ & $25\%$ & $50\%$ & $5\%$ & $15\%$ & $25\%$ & $50\%$ & $0\%$\tabularnewline
\hline 
\textbf{MNIST$\goto$MNIST-M} & 86.4 & 88.8 & 92.9 & \textbf{93.2} & 75.5 & 70.2 & 64.5 & \emph{58.4} & 81.5\tabularnewline
\hline 
\textbf{SVHN$\goto$MNIST} & 72.3 & 74.1 & 76.2 & \textbf{77.5} & 69.8 & 60.8 & 55.8 & \emph{56.4} & 71.0\tabularnewline
\hline 
\end{tabular}\caption{The variation of predictive performance in percentage as increasing
the ratio of labeled portion when the labels of the target domain
are properly or improperly matched to those in the source domain.
Note that we emphasize in bold and italic the best and worse performance.\label{tab:label_harmony}}
\vspace{-2mm}
\end{table}
It can be observed from the experimental results in Table \eqref{tab:label_harmony}
that for the case of proper matching, when increasing the ratio of
labeled portion, we increase the chance to match the corresponding
labeled portions properly, hence significantly improving the predictive
performance. In contrast, for the case of improper matching, when
increasing the ratio of labeled portion, we increase the chance to
match the corresponding labeled portions improperly, hence significantly
reducing the predictive performance.

%% file: Conclusion.tex
Deep domain adaptation is a recent powerful learning framework which
aims to address the problem of scarcity of qualified labeled data
for supervised learning. To enable transferring the learning across
the source and target domains, deep domain adaptation tries to bridge
the gap between the source and target distributions in a joint feature
space. Although this idea is powerful and has empirically demonstrated
its success in several recent work, its theoretical underpinnings
are lacking and limited. In this paper, we have developed a rigorous
theory to establish a firm theoretical foundation for deep domain
adaptation. Our theory provides a much more stronger theoretical results
with more realistic assumption for real-world applications compared
with existing work. Our result further offers a theoretical explanation
behind the rationale for deep domain adaptation approach in bridging
the gap between the source and target domains in a joint space. Interestingly,
our work provides a deep connection to unsupervised style transfer
where popular models such as CycleGAN and DiscoGAN can be rigorously
explained. Lastly, the merit of our work was further validated via
extensive experiments.